\documentclass{Styles/svproc}

\usepackage{url}
\usepackage{cite}
\usepackage{amsmath,amssymb,amsfonts}
\usepackage{algorithmic}
\usepackage{graphicx}
\usepackage{textcomp}
\usepackage{xcolor}

\begin{document}
    
    % start of a contribution
    \mainmatter
    
    \title{The Critical Role of Effective Communication in Human-Robot Collaborative Assembly}
    \titlerunning{Communication in HRC Assembly}  % abbreviated title (for running head)
    
    \author{Davide Ferrari \and Cristian Secchi}
    \authorrunning{Davide Ferrari et al.} % abbreviated author list (for running head)
    
    \tocauthor{Davide Ferrari and Cristian Secchi}
    
    \institute{Department of Sciences and Methods of Engineering, \\
        University of Modena and Reggio Emilia, Reggio Emilia, Italy,\\
        \email{\{davide.ferrari95, cristian.secchi\}@unimore.it},\\
        \texttt{https://dismi.unimore.it}
    }
    
    \maketitle
    
    \begin{abstract}
    
        In the rapidly evolving landscape of Human-Robot Collaboration (HRC), effective communication between humans and robots is crucial for complex task execution. Traditional request-response systems often lack naturalness and may hinder efficiency. This study emphasizes the importance of adopting human-like communication interactions to enable fluent vocal communication between human operators and robots simulating a collaborative human-robot industrial assembly.
        We propose a novel approach that employs human-like interactions through natural dialogue, enabling human operators to engage in vocal conversations with robots. Through a comparative experiment, we demonstrate the efficacy of our approach in enhancing task performance and collaboration efficiency. The robot’s ability to engage in meaningful vocal conversations enables it to seek clarification, provide status updates, and ask for assistance when required, leading to improved coordination and a smoother workflow. The results indicate that the adoption of human-like conversational interactions positively influences the human-robot collaborative dynamic. Human operators find it easier to convey complex instructions and preferences, resulting in a more productive and satisfying collaboration experience.
    
    \end{abstract}

    % Chapters Input
	\section{Introduction}\label{sec:introduction}

    The increasing integration of robots into industrial settings \cite{Vil18Mechatronics} has led to a more intricate landscape of collaboration between humans and machines \cite{Baratta2023PCS}. This new era of human-robot interaction (HRI) presents numerous opportunities to enhance work efficiency and productivity. However, it also requires a thoughtful examination of how to optimize communication processes among the involved participants \cite{Mukherjee2022Frontiers}. Effective communication between humans and robots, which mirrors the importance of communication in Human-Human Collaboration (HHC) \cite{Mojtahedi2017Frontiers}, represents a crucial factor in determining the success of these collaborative interactions.

    \noindent
    Current approaches often fall short of achieving the dynamic, bidirectional, and proactive communication characteristic of human interactions; traditionally, in a collaborative assembly scenario within an industrial environment, humans and robots have predefined tasks and interactions, with communication limited to occasional user requests utilizing a simple request-response mechanism \cite{MAVRIDIS201522}, providing direct commands to the robot. In \cite{Grushko2021Sensors} communication strategies have been employed to provide operators with insights into the intentions of robots to fosters coexistence and trust between humans and machines but also ensures that both parties remain well-informed about the planned actions of the robot. In \cite{Rosen2020IROS} a multimodal communication approach is used to enable unidirectional commands from the operator to the robot. While the communication in these cases is primarily unidirectional, it helps to reduce the potential for misunderstandings or unsafe actions. However, it's important to note that unidirectional communication has limitations, as it does not facilitate the exchange of feedback or convey essential information that can be crucial for promoting both safe and efficient collaboration between humans and robots. Conversely, in \cite{Ferrari2022ICRA}, voice communication is structured to empower the robot with the capability to initiate dialogues using a bidirectional voice communication framework, in order to let the robot communicate some problems and error that may occur during the collaborative job.

    \noindent
    While these approaches hold promise in enhancing human-robot communication, there are instances where relying solely on unidirectional or robot-initiated requests may prove to be limiting and inefficient, especially when confronted with complex and dynamic tasks. The main objective of this article is to implement a human-like \cite{Vargas2021Frontiers} natural and bidirectional vocal communication architecture within a collaborative human-robot assembly task involving an industrial component, developing more complex and natural communication and seeking to emulate what occurs in collaborative human-human assembly, leading to improvements in task performance and collaboration efficiency. The robot's capacity to engage in meaningful vocal conversations allows it to seek clarification, offer status updates, and request assistance when necessary, resulting in enhanced coordination and a smoother workflow.

    \noindent
    In the following sections of this paper, we will introduce our framework that incorporates natural communication within a collaborative assembly of an industrial component. Furthermore, we will conduct a comparative experiment to validate our proposal and illustrate how more sophisticated communication can bring benefits in collaboration and user experience. The contributions of this article encompass:
    \begin{itemize}
        \item An architecture for natural communication in a collaborative assembly task involving an industrial component.
        \item An experimental validation through a collaborative experiment to assess the validity of the proposed framework.
    \end{itemize}

    \section{Natural Collaborative Assembly Framework}\label{sec:architecture}

    Let's consider a scenario of Human-Robot Collaborative Assembly involving an industrial component, where both the robot and the human operator are engaged in a complex task that requires shared knowledge and mutual assistance. In this context, effective communication between the operator and the robot takes on primary importance. They need to be able to exchange information regarding task progress, request tools or components, seek assistance or support, and engage in dynamic, context-aware conversations. Building upon this, our proposed architecture aims to incorporate advanced natural language processing (NLP) techniques, leveraging pre-trained deep reinforcement learning models capable of generating a coherent and natural dialogue flow, thus emulating human capabilities. This will enable the robot to understand and respond to user requests in a more natural and context-aware manner.

\begin{figure}[htbp]
        \centering
        \includegraphics[trim={4cm 8cm 5cm 6.5cm}, clip, width=1.0\linewidth]{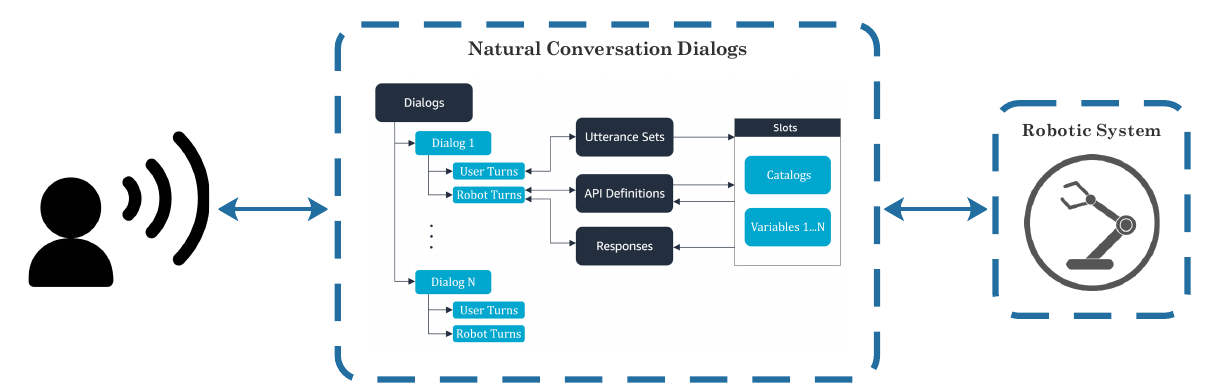}
        \caption{Natural Conversation Framework}
        \label{fig:Conversation-Structure}
    \end{figure}

    \vspace{\baselineskip}
    \noindent 
    To facilitate this natural communication, our proposed architecture incorporates a commercial voice assistant into the Human-Robot Collaboration job. This integration enables reciprocal information exchange between the robot and the operator through a voice communication channel structured in the form of conversations. Each conversation, as depicted in Figure \ref{fig:Conversation-Structure}, consists of a series of turn-taking dialogues, starting with the user's turn and then transitioning to the robot's turn. Within each dialogue, the user's requests are associated with predefined example phrases (utterance sets), which can be customized with variables or catalogs if required. The robot can respond directly, request additional information to fill in missing variable slots, or transmit the message to the back-end using a JSON-Request. Depending on the nature of the request, the system can call upon APIs that may reference external objects or trigger subsequent dialogues, allowing the conversation with the human to unfold in a contextually rich and meaningful manner. 
    
    \section{Experimental Validation}\label{sec:Experimental_Validation}

    The experimental validation\footnote{Detailed video of the experiment at: https://doi.org/10.5281/zenodo.10105470.} involved a comparative experiment that simulated a collaborative assembly task. In this scenario, an operator worked together a UR10e collaborative manipulator to assemble an industrial planetary gearbox. The components and tools required for the assembly were strategically positioned within the shared workspace, accessible to both the robot and the operator. Leveraging voice communication, the operator had the capability to request tools, components, or assistance, while the robot, through dialogues, could provide details, address issues, propose alternatives, and provide assistance. The experiment, carried out on a sample of 10 participants, randomizing the execution order of the two experimental setups to minimize the potential influence of a learning effect, aimed to compare the proposed conversation collaborative assembly architecture with the traditional industrial assembly job based on fixed task and a minimal request-response communication structure.

    \subsection{Implementation Details}\label{subsec:implementation}

        The architecture was implemented by integrating into the ROS framework
        % \cite{ros}
        a custom Amazon Alexa Conversations \cite{acharya2021alexa} skill, a deep learning-based approach that employs API calls to manage multi-turn dialogues between Alexa and the user, enabling more natural and human-like interactions. The skill's back-end was locally hosted, facilitating seamless integration with ROS by leveraging Microsoft Azure's HTTP Trigger Functions. Furthermore, to enable direct interaction with Alexa APIs, a Node-RED flow
        % \cite{nodered}
        was developed, a web service for logical path programming that allows event management and the initiation of conversations by invoking specific dialogue APIs.

   \subsection{Analysis of the Results}\label{subsec:results}

\begin{figure}[htbp]
            \centering
            \includegraphics[trim={0 0 0 0}, clip, width=0.805\linewidth]{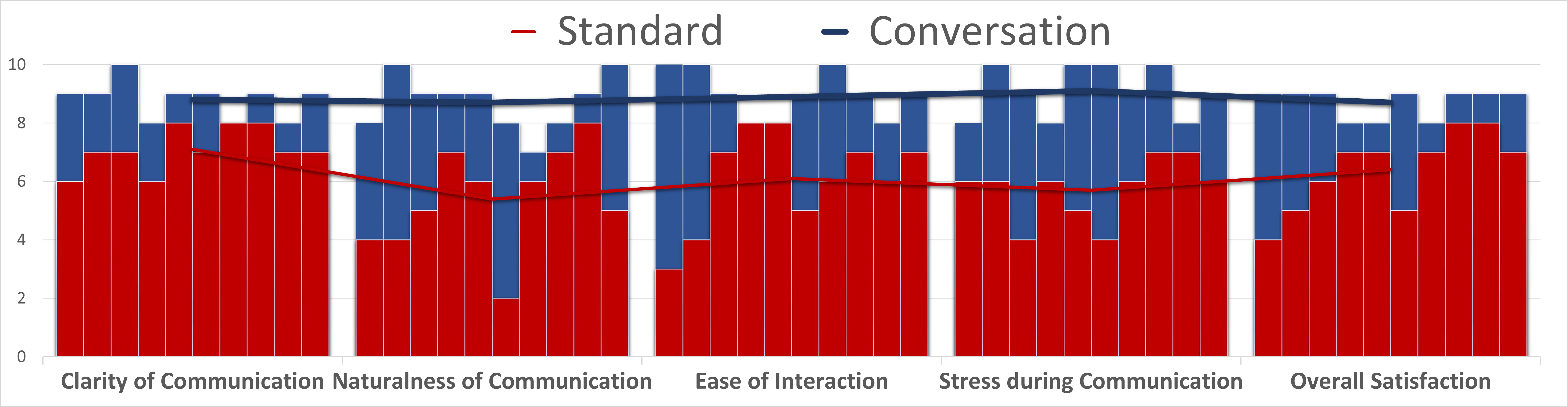}
            \begin{minipage}[b]{0.49\columnwidth}
                % \centering
                \raggedleft
                \includegraphics[trim={0 0 0 0}, clip, width=0.8\linewidth]{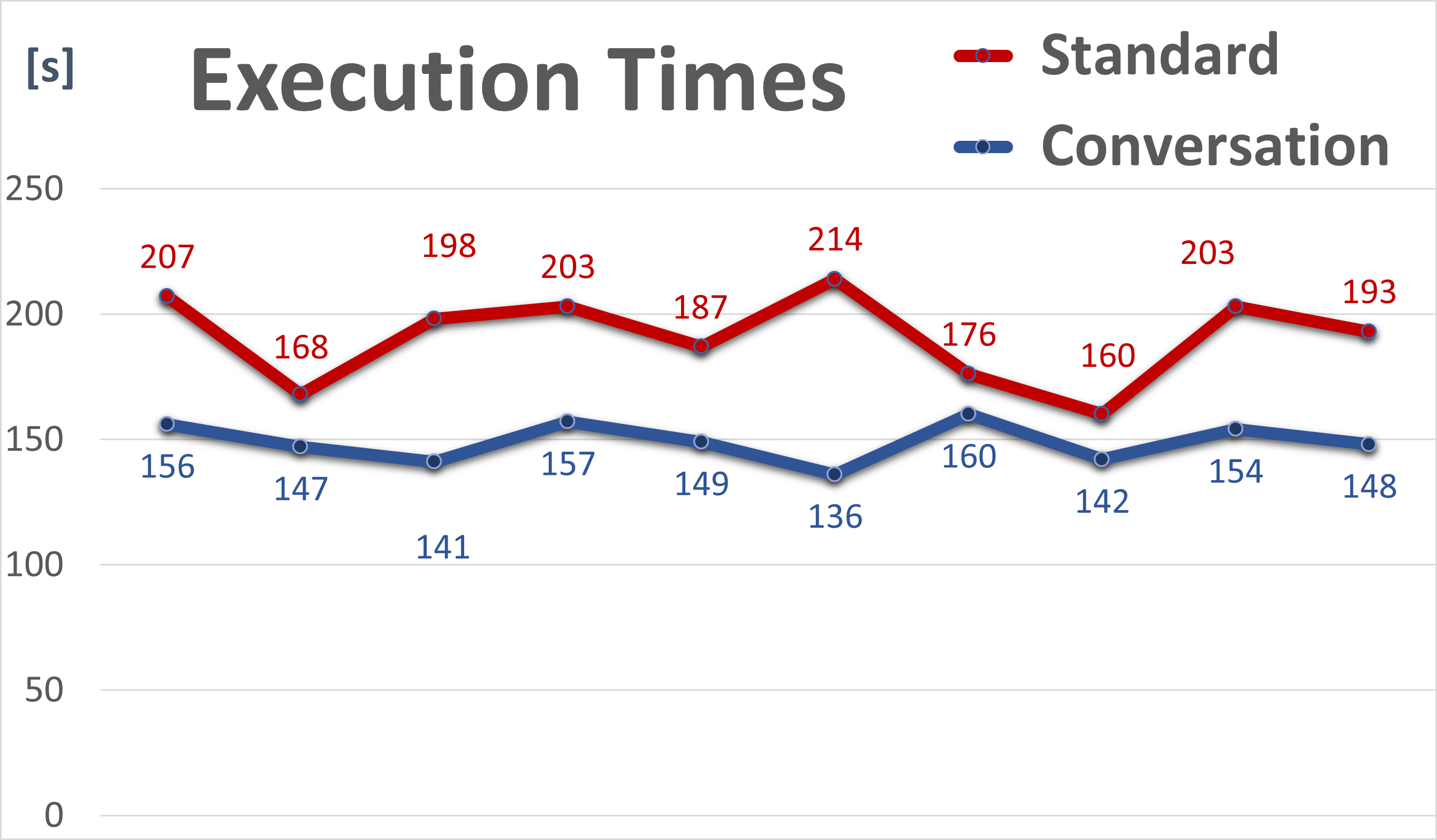}
            \end{minipage} %
            % \hfill
            \hspace{1pt}
            \begin{minipage}[b]{0.49\columnwidth}
                % \centering
                \raggedright
                \includegraphics[trim={0 0 0 0}, clip, width=0.8\linewidth]{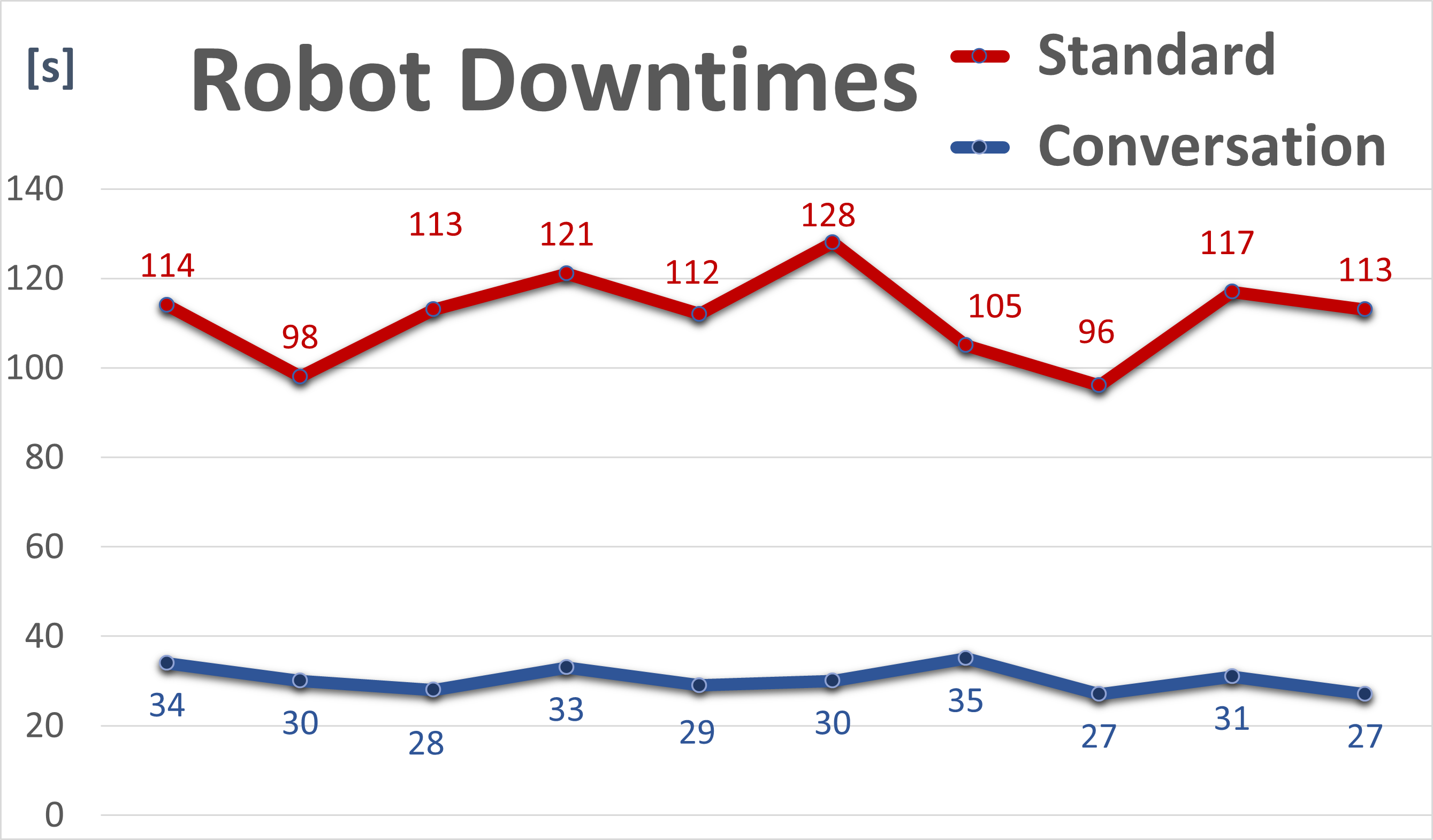}
            \end{minipage} \hfill
            \caption{Measured Times and Questionnaire Results}
            \label{fig:results}
        \end{figure}

        To evaluate the effectiveness of our architecture, we measured execution times and robot downtimes, and we collected the user feedback via a questionnaire consisting in five ratings on a scale from 0 to 10, covering key aspects: \textit{Clarity of Communication}, \textit{Naturalness of Communication}, \textit{Ease of Interaction}, \textit{Stress During Communication}, and \textit{Overall Satisfaction}.
        Figure \ref{fig:results} shows the results, with the comparative experiment in red and the proposed architecture in blue. The results underscore a notable difference: the proposed architecture received an average score of approximately 8.8/10, while the state-of-the-art approach averaged around 6.1/10. These findings indicate a significant enhancement in the user experience when utilizing our architecture, marked by improved clarity, ease of use, and reduced stress resulting from more natural communication. Moreover, the comparison of execution times and robot downtimes has demonstrated how the implementation of simple and natural communication within a collaborative assembly task can significantly reduce both execution times (22\%) and robot downtimes (73\%), resulting in increased efficiency and collaboration.

% 2 Dialogs:

%     Start the experiment.
%     Okay, to begin, take this part and mount the solar and planetary gears on it using the M4 screws. Do you need something else ?
%     Yes. Can you provide me with the screws ?
%     Certainly. Do you also need the screwdriver?
%     No, I can manage it on my own.

%     Help me by holding this.
%     Sure, I'm coming.
%     ...
%     Please hand me the object.
%     Here it is.
%     Should I move to the mounting position?
%     No, stay right here.
    
    % \bibliographystyle{Styles/bibtex/splncs03}
    \bibliographystyle{Styles/bibtex/splncs03_unsrt}
    \bibliography{bibliography}

\begin{thebibliography}{10}
\providecommand{\url}[1]{\texttt{#1}}
\providecommand{\urlprefix}{URL }

\bibitem{Vil18Mechatronics}
Villani, V., Pini, F., Leali, F., Secchi, C.: Survey on human–robot collaboration in industrial settings: Safety, intuitive interfaces and applications. Mechatronics  55,  248--266 (2018)

\bibitem{Baratta2023PCS}
Baratta, A., Cimino, A., Gnoni, M.G., Longo, F.: Human robot collaboration in industry 4.0: a literature review. Procedia Computer Science  217,  1887--1895 (2023)

\bibitem{Mukherjee2022Frontiers}
Mukherjee, D., Gupta, K., Najjaran, H.: A critical analysis of industrial human-robot communication and its quest for naturalness through the lens of complexity theory. Frontiers in Robotics and AI  9 (2022)

\bibitem{Mojtahedi2017Frontiers}
Mojtahedi, K., Whitsell, B., Artemiadis, P., Santello, M.: Communication and inference of intended movement direction during human–human physical interaction. Frontiers in Neurorobotics  11 (2017)

\bibitem{MAVRIDIS201522}
Mavridis, N.: A review of verbal and non-verbal human–robot interactive communication. Robotics and Autonomous Systems  63,  22--35 (2015)

\bibitem{Grushko2021Sensors}
Grushko, S., Vysocký, A., Oščádal, P., Vocetka, M., Novák, P., Bobovský, Z.: Improved mutual understanding for human-robot collaboration: Combining human-aware motion planning with haptic feedback devices for communicating planned trajectory. Sensors  21(11) (2021)

\bibitem{Rosen2020IROS}
Rosen, E., Whitney, D., Fishman, M., Ullman, D., Tellex, S.: Mixed reality as a bidirectional communication interface for human-robot interaction. In: 2020 IEEE/RSJ International Conference on Intelligent Robots and Systems (2020)

\bibitem{Ferrari2022ICRA}
Ferrari, D., Benzi, F., Secchi, C.: Bidirectional communication control for human-robot collaboration. In: 2022 International Conference on Robotics and Automation (ICRA). p. 7430–7436. IEEE Press (2022)

\bibitem{Vargas2021Frontiers}
Marin~Vargas, A., Cominelli, L., Dell’Orletta, F., Scilingo, E.P.: Verbal communication in robotics: A study on salient terms, research fields and trends in the last decades based on a computational linguistic analysis. Frontiers in Computer Science  2 (2021)

\bibitem{acharya2021alexa}
Acharya, A.: Alexa conversations: An extensible data-driven approach for building task-oriented dialogue systems (2021)

\end{thebibliography}

\end{document}